\documentclass[final]{cvpr}

\usepackage{times}
\usepackage{epsfig}
\usepackage{graphicx}
\usepackage{amsmath}
\usepackage{amssymb}
\usepackage{kotex}
\usepackage{booktabs}
\usepackage{multirow}
\usepackage{dsfont}
\usepackage[dvipsnames]{xcolor}
\usepackage{subcaption}
\usepackage{pifont}
\usepackage{floatrow}
\usepackage{nicefrac}
\usepackage{algorithm,algorithmicx,algpseudocode}
\usepackage[pagebackref=true,breaklinks=true,colorlinks,bookmarks=false]{hyperref}

\def\cvprPaperID{****} %
\def\confYear{****}

\begin{document}

\title{
VideoMix: Rethinking Data Augmentation for Video Classification
}

\author{
Sangdoo Yun$^1$, 
~Seong Joon Oh$^1$,
~Byeongho Heo$^1$, 
~Dongyoon Han$^1$, 
~Jinhyung Kim$^2$ \\ 
~ \\
NAVER AI LAB$^1$ \\
School of Electrical Engineering, KAIST$^2$ 
}

\maketitle

\begin{abstract}

State-of-the-art video action classifiers often suffer from overfitting. They tend to be biased towards specific objects and scene cues, rather than the foreground action content, leading to sub-optimal generalization performances. 
Recent data augmentation strategies have been reported to address the overfitting problems in static image classifiers. 
Despite the effectiveness on the static image classifiers, data augmentation has rarely been studied for videos. 
For the first time in the field, we systematically analyze the efficacy of various data augmentation strategies on the video classification task. We then propose a powerful augmentation strategy \textbf{VideoMix}.
VideoMix creates a new training video by inserting a video cuboid into another video. The ground truth labels are mixed proportionally to the number of voxels from each video. 
We show that VideoMix lets a model learn beyond the object and scene biases and extract more robust cues for action recognition.
VideoMix consistently outperforms other augmentation baselines on Kinetics and the challenging Something-Something-V2 benchmarks. It also improves the weakly-supervised action localization performance on THUMOS'14. VideoMix pretrained models exhibit improved accuracies on the video detection task (AVA).

\end{abstract}
\section{Introduction}
\label{section:introduction}
Video action classification models have achieved remarkable performance improvements in recent years. 
The main innovations have stemmed from the introduction of large-scale video dataset like Kinetics~\cite{kinetics} and Sports-1M~\cite{KarpathyCVPR14} and the development of powerful network architectures using 3D convolutional neural networks (3D CNNs). 

As the architecture of video models become deeper and more complex, overfitting and the resulting loss of generalizability become greater concerns. For example, models trained on large-scale video dataset still suffer from the object and scene biases: models rely heavily on specific discriminative objects and scene elements~\cite{sevilla2019only,li2019repair,weinzaepfel2019mimetics}. This has led to sub-optimal generalization performances and the decrease in localization abilities of video action classifiers. Ideally, a model should extract cues for recognition from diverse sources to enhance generalizability and the robustness to missing features.

The above problems are already identified and studied in static image recognition tasks, especially in the context of modern models based on 2D CNN architectures. For 2D image recognition, data augmentation has proven to be effective, while not requiring extra annotation or training time~\cite{singh2017hide,choe2019attention,cutmix}.

In contrast, there is a lack of extensive studies on the data augmentation strategies for video recognition tasks. In this paper, we examine the efficacy of image-domain data augmentation strategies on video data, especially the ones based on feature erasing that are known to improve model robustness and generalizability~\cite{devries2017cutout,zhong2017randomerase}.
In particular, we consider a generalization of CutMix~\cite{cutmix} to video sequences.
We show experiments to decide which axis (spatial or temporal) CutMix needs to be extended for the best performance on video sequences. 
As a result of our analysis, we introduce the \textbf{VideoMix} augmentation strategy. A new training video sample is constructed by cutting and pasting a random video cuboid patch from one video to the other. The ground truth label for this video is a volume-proportional combination of the source video labels.

\begin{table*}[t]
\centering
\tabcolsep=0.07cm
\begin{tabular}{@{}lccccc@{}}
\toprule
Task     & \multicolumn{3}{c}{Action recognition}                                                                                                                                             & Localization  & Detection     \\ \midrule
Dataset  & \begin{tabular}[c]{@{}c@{}}Kinetics-400\\ (acc. \%) \end{tabular} & \begin{tabular}[c]{@{}c@{}} Mini-Kinetics\\ (acc. \%) \end{tabular} & \begin{tabular}[c]{@{}c@{}} Something-V2\\ (acc. \%) \end{tabular} &  \begin{tabular}[c]{@{}c@{}} THUMOS'14\\ (mAP)  \end{tabular}       & \begin{tabular}[c]{@{}c@{}} AVA \\ (mAP)  \end{tabular}           \\ \midrule
Model    & SlowOnly-50                                               & SlowFast-50                                                  & SlowFast-50                                                        & I3D T-CAM         & SlowFast-50  \\ \midrule
Baseline & 73.6                                                   & 79.5                                                    & 61.5                                                            & 17.8          & 23.2          \\
VideoMix & \textbf{74.9}                                          & \textbf{81.9}                                           & \textbf{62.3}                                                   & \textbf{19.3} & \textbf{24.9} \\
Improve. $\Delta$ & ($\mathbf{+1.3}$) & ($\mathbf{+2.4}$) & ($\mathbf{+0.8}$) & ($\mathbf{+1.5}$) & ($\mathbf{+1.7}$) \\ \midrule
\end{tabular}
\caption{\textbf{Overview of VideoMix performances.} 
We compare our VideoMix against the vanilla training strategy (Baseline) on various tasks.
VideoMix consistently improves action recognition, localization, and detection performances without any added parameter or a significant amount of computational overhead.
}
\label{table:intro}
\end{table*}

We show the effectiveness of VideoMix with extensive evaluations on various 3D CNN architectures, datasets, and tasks.
Table~\ref{table:intro} summarizes the improvements by VideoMix.
VideoMix consistently improves video classification models without any additional parameter and a significant amount of computation.

\section{Related Works}
\label{section:related_works}

We briefly discuss the related works of the video classification task and the data augmentation in this section. 

\subsection{Video Classification}

The key difference between video and image classifications is that the former must capture temporal information.
Some prior works~\cite{simonyan2014two,feichtenhofer2016convolutional,carreira2017quo} have extracted temporal cues explicitly (e.g. via optical flow).
With the development of deep learning and large-scale video datasets~\cite{kinetics,sports1M}, 3D convolutional neural networks (3D CNNs)~\cite{tran2015learning_spatiotemporal,hara2018can_spatiotemporal,carreira2017quo,tran2019CSN} and non-local modules~\cite{wang2018non}  
are proposed to learn the temporal cues automatically.
Recently proposed SlowFast network~\cite{feichtenhofer2019slowfast} and CSN~\cite{tran2019CSN} show the state-of-the-art performances on the video classification using 3D CNNs. 
SlowFast proposed a dual-branch architecture to combine a slow path for static spatial features and a fast path for dynamic motion features, and CSN utilizes depth-wise convolution for lightweight 3D CNN architecture. 
While the advances in video classification have been focused on the architectural axis, we explore the orthogonal \textbf{data} axis, which is seldom explored in the context of video recognition tasks. 
We show the effectiveness of VideoMix by conducting experiments on top of the state-of-the-art SlowFast and CSN networks.

\subsection{Data Augmentation}

\paragraph{Data augmentation for image classification.}
There are many augmentation strategies for static image classification tasks. 
Horizontal flipping, random resizing, and cropping have been used in training ImageNet classifiers and are now considered the standard set of augmentation strategies~\cite{Inceptionv3}.
There have been regional dropout methods~\cite{devries2017cutout,zhong2017randomerase} which remove random regions of an image to enhance robustness and generalization.
Other single-image augmentation strategies include RandAugment~\cite{randaugment} and AutoAugment~\cite{autoaugment}. They consider the combination of extensive pixel-level image augmentation types, such as rotation, shear, translation, and color jittering. RandAugment and AutoAugment train classifiers with above operations via random selection and learned policy, respectively.
Augmentation strategies that combine more than one image include Mixup~\cite{zhang2017mixup} and CutMix~\cite{cutmix}. Mixup averages the RGB values of two images and the ground truth labels to create new samples.
CutMix~\cite{cutmix} has improved upon regional dropout by filling in image patches from other images in the dropped-out region, thereby maximizing the pixel efficiency during training.
The labels are mixed among the source images as in Mixup. We discuss and experiment with the above static-image augmentation strategies on video classification tasks in our analysis.

\paragraph{Data augmentation for video classification.}
There do exist a few attempts to apply data augmentation strategies on videos. On the spatial side, the standard single-image augmentation strategies for image classification have been considered: horizontal flipping and random cropping~\cite{wang2015towards}. 
Along the temporal axis, a widely-used augmentation strategy is to randomly sub-sample a shorter video clip from the full sequence.
However, there has been an overall lack of extensive studies on video augmentation methods. This work contributes the first studies on the impact of video augmentation strategies on the generalization, localization, and transfer learning capabilities.
\section{VideoMix}
\label{section:videomix}

In this section, we analyze existing data augmentations, describe the VideoMix algorithm, and discuss the effectiveness of VideoMix. 

\subsection{Revisiting Data Augmentation Techniques}

\begin{table}[t]
\centering
\begin{tabular}{@{}lccc@{}}
\toprule
Methods              & top1  & top5  \\ \midrule
Vanilla      &   75.2 &  91.7    \\
Mixup~\cite{zhang2017mixup}         &   77.0 &  93.1     \\
RandAugment~\cite{randaugment}   &   75.6 &   92.2    \\ 
Cutout~\cite{devries2017cutout}        &   76.1 &  92.6      \\
CutMix~\cite{cutmix}      &  76.7    &  92.9     \\  
\midrule
VideoMix       &  \textbf{77.6}    &  \textbf{93.5} \\
\midrule
\end{tabular}
\caption{\textbf{VideoMix vs existing data augmentation techniques.} The compared baselines are originally proposed for the image classification tasks. We extend them from two spatial dimensions to three spatio-temporal dimensions. SlowOnly-34 ($8\times8$) network and Mini-Kinetics dataset are used. 
}
\label{table:method:analysis}
\end{table}

We review existing data augmentation methods that partially remove regions in an image~\cite{devries2017cutout}, add pixel-level noises~\cite{autoaugment,randaugment}, or manipulate both input pixels and labels~\cite{zhang2017mixup,cutmix}.
We first evaluate their effectiveness by simply extending the two-dimensional methods to the three spatio-temporal dimensions.

Table~\ref{table:method:analysis} compares the augmentation strategies including Mixup~\cite{zhang2017mixup}, RandAugment~\cite{randaugment}, Cutout~\cite{devries2017cutout}, and CutMix~\cite{cutmix}.
It is straightforward to extend Mixup and RandAugment to videos: they apply global operations over the pixels. Mixup averages the RGB values of two images and RandAugment applies rotation, shear, and uniform perturbations on the images. We apply the same operation over the spatio-temporal frames.
For Cutout and CutMix, we choose the full spatio-temporal extension where sub-cuboids in videos are randomly selected to be removed or replaced.
We set the hyperparameter $\alpha$ of Mixup and CutMix to $1.0$, the mask size for Cutout to $112\times112$, the magnitudes $M$ for RandAugment to $9$.
Table~\ref{table:method:analysis} shows that even the naive extension of the baselines lead to improvements in video classification against the vanilla model. For example, Mixup achieves 77.0\% top-1 accuracy on Mini-Kinetics, an $+\mathbf{1.8}\%$ improvement against the vanilla SlowOnly-34 ($8\times8$) model.

In the rest of the section, we seek ways to boost the video recognition performances further by studying the design choices in augmentation strategies in greater depth.

\subsection{VideoMix Algorithm}
\label{section:videomix:algorithm}
We introduce the \textbf{VideoMix} algorithm.
Let $x \in \mathbb{R}^{T \times W \times H}$ be a video sequence, where $T$, $W$, and $H$ are the number of frames, width, and height of video sequences, respectively\footnote{We omit the input channel dimension (3 for RGB) for simplicity.}. Let $y$ be the corresponding ground truth label, represented as a one-hot vector. 
VideoMix generates a new training sample $(\Tilde{x},\Tilde{y})$ by combining two training samples $(x_{A}, y_{A})$ and $(x_{B}, y_{B})$. 
The generated $(\Tilde{x},\Tilde{y})$ is used for training the model with its original loss function.

More precisely, VideoMix first defines a binary tensor mask $\mathbf{M} \in \{0,1\}^{T \times W \times H}$ signifying the cut-and-paste locations in two video tensors. The new video is generated by the below procedure:
\begin{equation}
\begin{split}
    \Tilde{x} & =  \mathbf{M} \odot x_{A} + (\mathbf{1}- \mathbf{M}) \odot x_{B} \\
    \Tilde{y} & =  \lambda_\mathbf{M} y_A + (1-\lambda_\mathbf{M}) y_B
\end{split}
\label{eq:cutmix}
\end{equation}
where $\odot$ is the element-wise multiplication and $\lambda_\mathbf{M} := \frac{1}{TWH}\sum_{t,w,h}\mathbf{M}_{t,w,h}$
denotes the proportion of the volume occupied by $\mathbf{M}$.

The binary tensor mask $\mathbf{M}$ is decided by drawing the 3D cuboid coordinates $\mathbf{C}=(t_1,t_2,w_1,w_2,h_1,h_2)$ at random. More specifically,
\begin{align}
\small
    \mathbf{M}_{t,w,h}:=
    \begin{cases}
        1,&\text{ if }\quad t_1\leq t \leq t_2\text{, }w_1\leq w \leq w_2\text{, and } h_1\leq h \leq h_2 \\
        0,&\text{ otherwise}
    \end{cases}
\end{align}
We will investigate the design choices for the random selection of coordinates in the next part.

\subsection{Investigation of spatial and temporal axes for VideoMix}
\begin{figure}[!t]
\centering
\includegraphics[width=\linewidth]{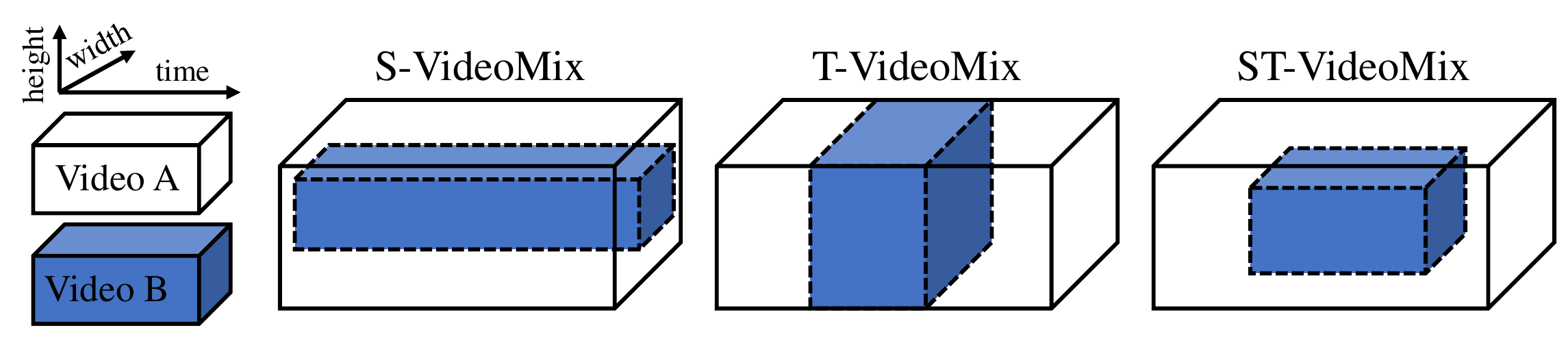}
\caption{\textbf{VideoMix variants.} Illustrations of Spatial (S-), Temporal (T-), and Spatio-temporal (ST-) VideoMix. 
We omit the channel (color) dimension for clarity. A sub-cuboid of Video B (blue cube) is inserted into Video A (white cube).}
\label{fig:videomix:videomix_types}
\end{figure}
\begin{table}[!t]
\centering
\begin{tabular}{@{}lccc@{}}
\toprule
Methods              & top1  & top5  \\ \midrule
Spatial VideoMix       &  \textbf{77.6}    &  \textbf{93.5} \\
Temporal VideoMix       &   75.6   & 92.5 \\
Spatio-temporal VideoMix       & 76.7  & 92.9\\
\midrule
\end{tabular}
\caption{\textbf{Performances of VideoMix variants.} Performances of Spatial, Temporal, and Spatio-temporal VideoMix. SlowOnly-34 ($8\times8$) network and Mini-Kinetics dataset are used.
}
\label{table:method:videomix_types}
\end{table}

We identify three types of VideoMix: Temporal, Spatial, and Spatio-temporal VideoMix.
Temporal VideoMix samples the cuboid coordinates $\mathbf{C}$ only along the temporal axis ($(t_1,t_2)$ are sampled) and fixes spatial coordinates at $(w_1,w_2,h_0,h_1)=(0,W,0,H)$. Spatial VideoMix samples the spatial coordinates $(w_1,w_2,h_1,h_2)$, while fixing $(t_1,t_2)=(0,T)$. 
Spatio-temporal VideoMix samples all the coordinates $(t_1,t_2,w_1,w_2,h_1,h_2)$.
The VideoMix variants are illustrated in Figure~\ref{fig:videomix:videomix_types}. 
Note that the Spatio-temporal VideoMix is the same as the CutMix in Table~\ref{table:method:analysis}.

Table~\ref{table:method:videomix_types} compares the performances of the VideoMix variants. Spatial VideoMix is the best among them. We hypothesize that the video sub-cuboid must secure a sufficient number of frames to represent the semantic information for the video category. Temporal VideoMix or Spatio-temporal VideoMix is limited in terms of the semantic content in cut and pasted video cuboids.
Spatial VideoMix, on the contrary, retains the full temporal semantics.

Based on this observation, we define \textbf{Spatial VideoMix} as our default VideoMix setting for video classification task.
The random coordinate selection strategy follows that of CutMix~\cite{cutmix}.
The spatial ratio $\lambda$ is sampled from the Beta distribution Beta($\alpha$,$\alpha$), where we set $\alpha:=1$\footnote{When $\alpha=1$, it is the uniform distribution $\text{Unif}(0,1)$.}.
The center coordinates $(w_c,h_c)$ are sampled from $\text{Unif}(0,W)\times\text{Unif}(0,H)$. Other cuboid endpoints are determined by 
\begin{equation}
\begin{split}
    w_{1} = w_c - \frac{W\sqrt{\lambda}}{2}~~~ & w_{2} = w_c + \frac{W\sqrt{\lambda}}{2} \\
    h_{1} = h_c - \frac{H\sqrt{\lambda}}{2}~~~ & h_{2} = h_c + \frac{H\sqrt{\lambda}}{2}
\end{split}
\label{eq:sampling}
\end{equation}
and fixed $(t_1,t_2)=(0,T)$.
Codes for VideoMix variants are presented in Appendix~\ref{appendix:algorithm}. 

\paragraph{VideoMix for temporal localization.}
Under certain application scenarios, it is difficult to directly manipulate the input videos. For example, features may have already been extracted from the original frames and the raw frames are unavailable because of the storage limits or legal issues~\cite{abu2016youtube}. 
As we will cover in the experiments of temporal action localization, \textbf{Temporal VideoMix} is a good alternative that improves the localization ability of a video classifier.

\begin{figure}[!t]
\centering
\includegraphics[width=\linewidth]{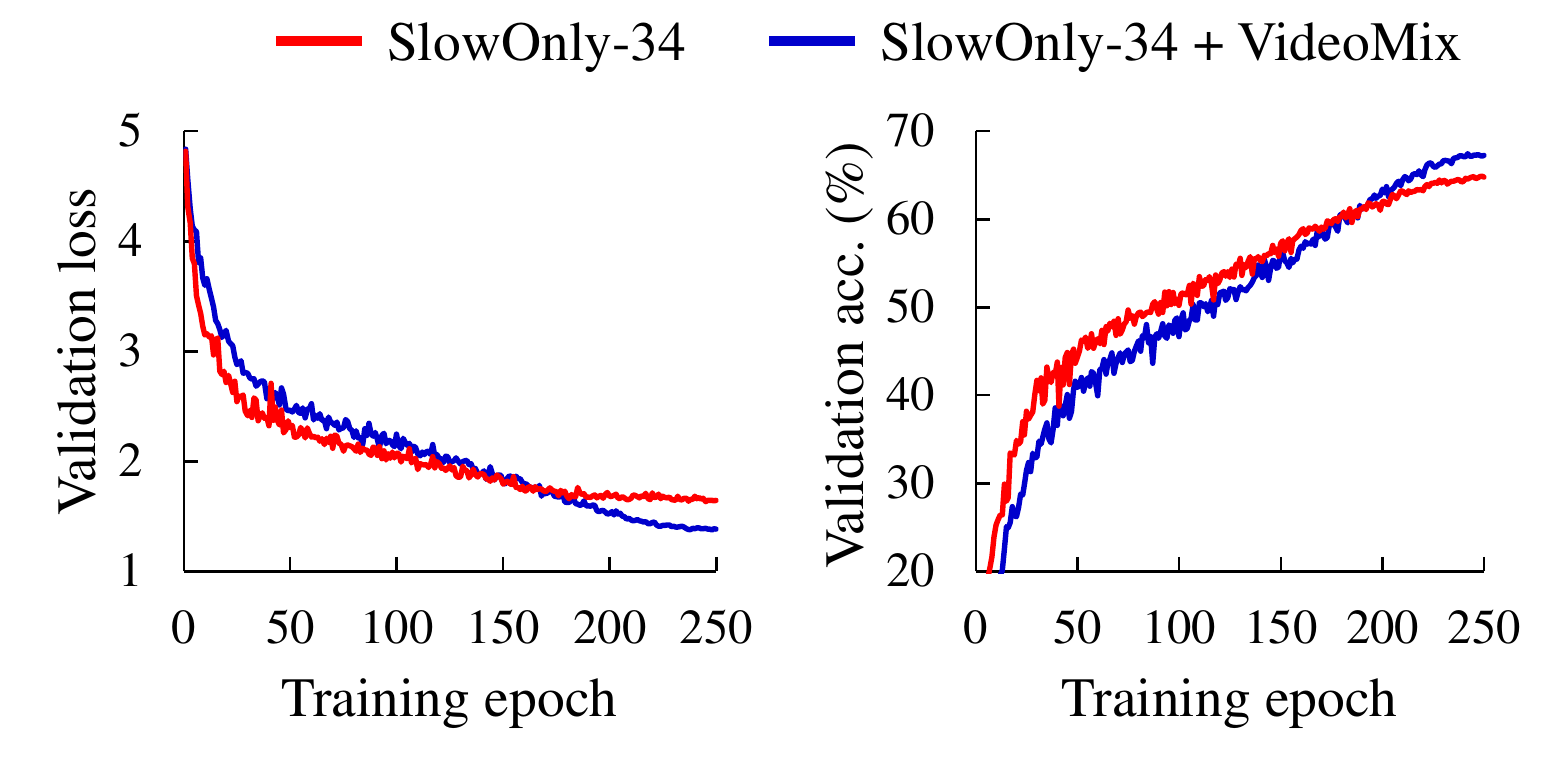}
\caption{\textbf{Training curves.} Validation loss (left) and top-1 accuracy (right) plots for Mini-Kinetics action recognition. VideoMix achieves lower validation loss and higher top-1 accuracy than the vanilla model.}
\label{fig:videomix:loss_plot}
\end{figure}

\subsection{Discussion}

\paragraph{Effect on preventing overfitting.}

To see the effect of VideoMix on preventing overfitting and stabilizing the training process, we compare validation loss and validation accuracy of SlowOnly-34 models during the training over Mini-Kinetics action recognition dataset in 
Figure~\ref{fig:videomix:loss_plot}.
We confirm that VideoMix enables video models to obtain lower validation loss and higher validation accuracy than the baseline. 
After about 200 training epochs, the baseline performance saturates and the loss does not decrease further.
Applying VideoMix lets the model overcome the barrier and improve further beyond this point. The training samples generated by VideoMix allow the video classifiers to generalize better.

\begin{figure*}[!t]
\centering
\includegraphics[width=0.80\linewidth]{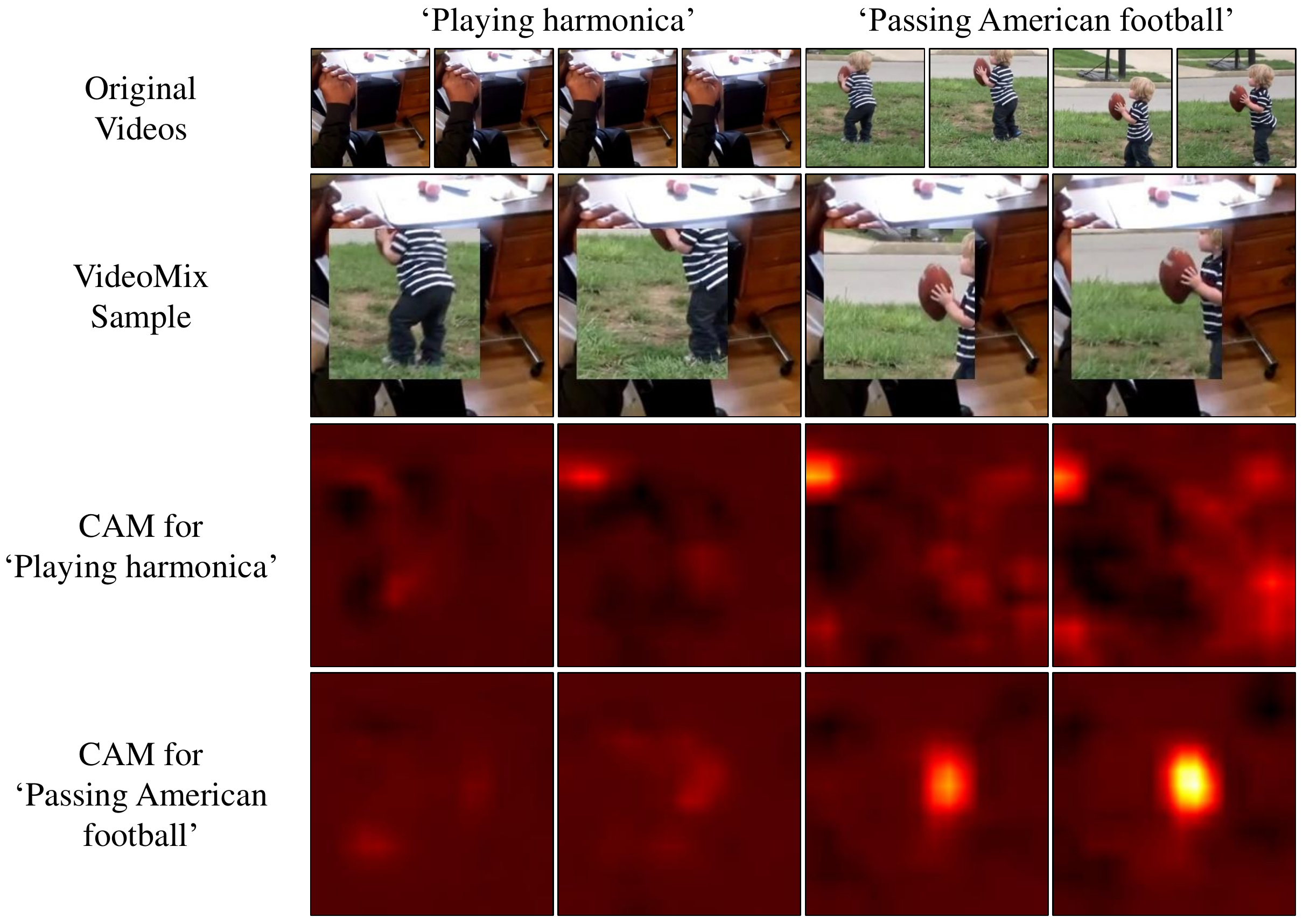}
\caption{\textbf{Class activation mapping (CAM) on VideoMix sample.} We show the spatio-temporal CAM~\cite{zhou2016CAM} score maps on a VideoMix sample. It is a combination of the ``playing harmonica'' and ``passing American football'' videos. The CAM score maps are visualized with respect to the two action classes separately. 
}
\label{fig:videomix:cam_visualize}
\end{figure*}

\paragraph{What does the model learn with VideoMix?}

We expect VideoMix to let an action classifier simultaneously recognize multiple actions present in the mixed videos. To verify that this is achieved, we visualize the spatio-temporal attention of a video on the synthetic video generated by VideoMix with the class activation maps (CAM)~\cite{zhou2016CAM}.
We use a Kinetics-400 pre-trained SlowOnly-50 model. 
We extend the original CAM proposed for static images to its spatio-temporal version that generates the $T$ sequential score maps. 
Detail description of spatio-temporal CAM and more examples are in Appendix~\ref{appendix:algorithm}. 
Figure~\ref{fig:videomix:cam_visualize} shows VideoMix samples and corresponding class activation maps with respect to the two action classes, ``playing harmonica'' and ``passing American football''.
The CAM results show that VideoMix guides a model to see multiple actions at once (e.g., the CAM for ``playing harmonica'' highlights the player's mouth and hands, and the CAM for ``passing American football'' emphasizes the kid's hands and the football object).
Furthermore, VideoMix reduces scene contexts of videos, as the background scene of ``passing American football'' are partially removed, and also hides some object cues, as the football 
of ``passing American football'' and the hands holding a harmonica of ``playing harmonica'' are blocked in some frames, which leads to learning more robust and generalized cues beyond the object and scene for action recognition.

\section{Experiments}
\label{section:experiments}

We evaluate VideoMix in terms of the improved generalization performances as well as the transfer-learning performances of VideoMix pre-trained models on multiple tasks.
We first verify the effect of VideoMix on video classification tasks: Kinetics action recognition~\cite{kinetics} and Something-Something-V2. 
We show the temporal action localization performance via weakly-supervised temporal action localization experiments.
Finally, VideoMix is evaluated in terms of the transfer-learning performances on the video action detection task.
We also provide additional experiments on HMDB-51~\cite{kuehne2011hmdb} and UCF-101~\cite{soomro2012ucf101} action recognition benchmarks in Appendix~\ref{appendix:more_baseline}.  
All experiments were implemented and evaluated on NAVER Smart Machine Learning (NSML)~\cite{nsml} platform
with PyTorch~\cite{paszke2017automatic}

\subsection{Kinetics Action Recognition}
\label{section:experiments:kinetics}

\paragraph{Dataset.}
Kinetics-400~\cite{kinetics} is a widely used large-scale action recognition benchmark consisting of 240k training videos and 20k validation videos in 400 human action classes.
The performances are evaluated with the top-1 and top-5 accuracies. 
Note that about 10\% of the Kinetics-400 videos are not available on YouTube to be downloaded. It is not possible to reproduce the exact accuracies reported in the original paper; we train the action classifier over the available subset and treat this result as the baseline.

\paragraph{Training and evaluation.}
We follow the original training recipes of the baseline architectures in~\cite{feichtenhofer2019slowfast}.
We train models from scratch using the stochastic gradient descent optimizer for 250 epochs with batch size $64$ and initial learning rate $0.1$, which is decayed by cosine annealing. 
For a training video clip, $64$ consecutive frames are randomly sampled from a video and $T$ frames are sub-sampled with $\tau$ temporal stride as an input for video models. 
For every model, random resize crop and random horizontal flip are applied on training video clips as standard augmentations.
For evaluation, we use $3\times10$ view ensembles as in~\cite{feichtenhofer2019slowfast}, where $10$ clips are uniformly sampled along the temporal dimension from the entire video sequences and $3$ spatial regions are uniformly sampled along the longer side of the frames.

\paragraph{Network architecture.}
We use the SlowOnly, SlowFast~\cite{feichtenhofer2019slowfast}, and interaction preserved CSN (ip-CSN)~\cite{tran2019CSN} to show the impact of VideoMix on the video classification task. 
Every model is based on the ResNet architecture~\cite{resnet}. We denote the specific ResNet type with the suffix ``-(\#\ignorespaces depth)''.
We also denote each video model with frame length $T$ and temporal stride $\tau$ in the trailing bracket $(T\times\tau)$.
For example, SlowOnly-50 ($4\times16$) is based on the ResNet-50 architecture, considers $T=4$ input frames sub-sampled from the original $64$ frames with temporal stride $\tau=16$.
SlowFast-50 ($8\times8$) takes two separate input streams, the slow and fast pathways, each with $8$ and $32$ total number of input frames ($T$) with temporal strides ($\tau$) $8$ and $2$, respectively.

\paragraph{Kinetics-400 results.}

\begin{table}[t]
\small
\tabcolsep=0.08cm
\centering
\begin{tabular}{@{}lcccc@{}}
\toprule
{Model}                     & VideoMix & ~~top1~  & ~top5~~    & GFlops$\times$views \\ \midrule
I3D              &       &  72.1     &  90.3  &   108 $\times$ N/A        \\
Two-Stream I3D   &       &  75.7     &  92.0  &  216  $\times$ N/A      \\
Nonlocal-ResNet50   &       &   76.5     &  92.6  &  282 $\times$ 30         \\
\midrule
SlowOnly-50 ($4\times16$)   &      &    71.8  & 89.6  & 26 $\times$ 30 \\ 
SlowOnly-50 ($4\times16$)   & \checkmark     &    \textbf{72.7}  & \textbf{90.3} & 26 $\times$ 30  \\
\midrule
SlowOnly-50 ($8\times8$)    &      &    73.6  & 90.7  & 54 $\times$ 30 \\
SlowOnly-50 ($8\times8$)    & \checkmark     &    \textbf{74.9}  & \textbf{91.7} & 54 $\times$ 30 \\
\midrule
SlowFast-50 ($8\times8$)    &       &   75.9  & 91.9 & 65 $\times$ 30 \\
SlowFast-50 ($8\times8$)    & \checkmark     &    \textbf{76.6}  & \textbf{92.6} & 65 $\times$ 30 \\ \midrule
\end{tabular}
\caption{\textbf{Kinetics-400 action recognition results.} The inference cost is reported in the last column with GFlops of a single view $\times$ the number of views. 
N/A indicates the number of views are not available for us.}
\label{table:experiment:kinetics400}
\end{table}

We evaluate VideoMix on Kinetics-400 with SlowOnly and SlowFast networks~\cite{feichtenhofer2019slowfast} as the base network architectures.
SlowFast combines two branches: the slow branch for static spatial features and the fast branch for dynamic motion features.
SlowOnly only has the slow branch that is similar to the ResNet~\cite{resnet} architecture with 3D convolutional kernels.
The experimental results are shown in Table~\ref{table:experiment:kinetics400}.
The accuracies in the table are reproduced results\footnote{The original paper has reported the top-1 accuracies for {SlowOnly-50 ($4 \times 16$), SlowOnly-50 ($8 \times 8$), and SlowFast-50 ($8 \times 8$)} as $72.6$, $74.8$, and $77.0$, respectively. The difference is due to the 10\% unavailable videos in Kinetics-400 and smaller batch sizes due to GPU limitations.}.
We also report the inference cost (GFlops) of a single view (a temporal clip with spatial crop) as well as the number of views for the prediction of a single video.
We observe that VideoMix consistently improves the accuracy of baseline models. 
VideoMix achieves the top-1 accuracy of $72.7\%$, $74.9\%$, and $76.6\%$ for SlowOnly-50 ($4 \times 16$), SlowOnly-50 ($8 \times 8$), and SlowFast-50 ($8 \times 8$) with improvements of $+\mathbf{0.9}\%$, $+\mathbf{1.3}\%$, and $+\mathbf{0.7}\%$, respectively.
We also show that VideoMix-augmented SlowFast recognizer achieves a competitive performance ($76.6\%$) against other methods such as I3D~\cite{carreira2017quo} ($72.1\%$), Two-Stream I3D~\cite{carreira2017quo} ($75.7\%$), and Nonlocal-ResNet50~\cite{wang2018non} ($76.5\%$) which require more computational costs (GFlops) than the SlowFast architecture. 

\paragraph{Mini-Kinetics results.}

\begin{table}[t]
\centering
\begin{tabular}{@{}lcccc@{}}
\toprule
{Model}     & VideoMix        & top1  & top5 \\ 
\midrule
ip-CSN-50 ($8\times8$)     &                 &  74.8    &  91.9     \\
ip-CSN-50 ($8\times8$)     & \checkmark    & \textbf{75.9} & \textbf{93.1}    \\
\midrule
SlowOnly-50 ($4\times16$)     &             &   74.4    &   91.3    \\
SlowOnly-50 ($4\times16$)&  \checkmark               & \textbf{76.0} & \textbf{93.0}       \\
\midrule
SlowOnly-50 ($8\times8$)     &             &   77.5    &   93.2    \\
SlowOnly-50 ($8\times8$)     & \checkmark  & \textbf{79.2} & \textbf{94.1}       \\
\midrule
SlowFast-50 ($8\times8$)     &             &    79.5   & 93.9      \\
SlowFast-50 ($8\times8$)     & \checkmark  &    \textbf{81.9}  &  \textbf{95.1}      \\ 
\midrule
\end{tabular}
\caption{\textbf{Mini-Kinetics action recognition results.}
}
\label{table:experiment:mini-kinetics}
\end{table}

We also evaluate VideoMix on Mini-Kinetics as shown in Table~\ref{table:experiment:mini-kinetics}.
We observe that VideoMix improves the performance of various baseline architectures: ip-CSN-50~\cite{tran2019CSN}, SlowOnly-50 ($4\times16$), SlowOnly-50 ($8\times8$), and SlowOnly-50 ($8\times8$) with $\mathbf{75.9}\%$ ($\mathbf{+1.1}\%$), $\mathbf{76.0}\%$ ($\mathbf{+1.6}\%$), $\mathbf{79.2}\%$ ($\mathbf{+1.7}\%$) and $\mathbf{81.9}\%$ ($\mathbf{+2.4}\%$), respectively.

\paragraph{Ablation Studies.}

We conduct ablation studies on the Mini-Kinetics dataset.
SlowOnly-34 is used as the running baseline, where the BasicBlock is used as in ResNet-34~\cite{resnet}.
The results are shown in Table~\ref{table:experiment:albation}.

\begin{table}[t]
\centering
\begin{tabular}{@{}lccc@{}}
\toprule
Augmentation type         & top1 & top5 \\ \midrule
Baseline    &  75.2   & 91.7 \\
VideoMix (Ours; $\alpha$=1.0) &  \textbf{77.6}   & {93.5} \\ \midrule
VideoMix ($\alpha$=0.2)     &  77.0   & {93.5} \\
VideoMix ($\alpha$=0.5)     &  \textbf{77.6}   & 93.4 \\
VideoMix ($\alpha$=2.0)     &  77.3   & \textbf{93.6} \\
VideoMix ($prob$=0.5)          &  77.0   & 93.0 \\\midrule
VideoMix (\#videos=3)     &  75.7   & 93.0 \\
VideoMix (\#videos=4)     &  71.9   & 91.4 \\ \midrule
Temporal VideoMix           &  75.6   & 92.5 \\
Spatio-temporal VideoMix    &  76.7  & 92.9 \\
Per-frame VideoMix          &  74.8   & 92.8  \\  
\midrule
\end{tabular}
\caption{\textbf{Ablation studies.} Results are based on SlowOnly-34 ($8\times8$) and Mini-Kinetics.}
\label{table:experiment:albation}
\end{table}

We first examine the impact of the mixture area hyperparameter $\alpha \in \{0.2,0.5,$ $1.0\text{(ours)},2.0\}$.
VideoMix at various $\alpha$ values exhibits stable performances, uniformly outperforming the vanilla baseline. VideoMix is not sensitive to the hyperparameter $\alpha$. 
When we reduce the chance of applying VideoMix on a minibatch to $prob$=$0.5$ (default is $prob$=$1.0$), the top-1 accuracy drops by $0.6$ percent point. The simple strategy of applying VideoMix on every video in the batch is a better choice.
When we increase the number of mixing videos (``\#videos") more than two, the performance significantly drops, which indicates that two videos are enough for VideoMix and mixing more than two videos may hinder the training convergence.

Temporal VideoMix and Spatio-temporal VideoMix are not as effective as the VideoMix (Spatial VideoMix), with only $+0.4\%$ and $+1.5\%$ boosts over the baseline.
Per-frame VideoMix independently applies VideoMix at every frame, ruining the temporal continuity of the VideoMix operation. It shows an even worse accuracy of 74.8\%, lower than the original model. Temporal continuity is an important ingredient for the video augmentation.

More ablation studies to see the effect of dataset sizes and applying other data augmentation methods together with VideoMix are provided in Appendix~\ref{appendix:more_baseline}.

\subsection{Something-Something-V2}
\label{section:experiments:something}

\begin{table}[t]
\centering
\begin{tabular}{@{}lcccc@{}}
\toprule
{Model}                & VideoMix      & top1 & top5   \\ \midrule
SlowOnly-50 ($8\times8$)         &        &  59.1   & 85.1 \\
SlowOnly-50 ($8\times8$) & \checkmark    &  \textbf{60.0}   & \textbf{86.0} \\  \midrule
SlowFast-50 ($8\times8$) &  &   61.5   & 86.9  \\
SlowFast-50 ($8\times8$)    &   \checkmark    & \textbf{62.3}     & \textbf{87.6}  \\  \midrule
\end{tabular}
\caption{\textbf{Something-Something-V2 action recognition results.}
}
\label{table:experiment:something}
\end{table}

\paragraph{Dataset.}
Something-Something-V2 dataset~\cite{goyal2017something} contains 169k training and 25k validation videos with 174 action classes.
We evaluate the performances with top-1 and top-5 accuracies. 
Something-Something-V2 is known for the fine-grainedness of actions, the diversity of contexts. It poses new challenges for action recognition not covered by Kinetics~\cite{kinetics}.

\paragraph{Implementation details.}
We use SlowOnly-50 ($8\times8$) and SlowFast-50 ($8\times8$) models.
The models are pre-trained on Kinetics-400 with the standard training strategy, and the final fully-connected layer is replaced with the new one with $174$ output dimensions. 
The entire models are then fine-tuned for the Something-Something-V2 dataset for $40$ epochs with the batch size $64$ and learning rate $0.01$, which is decayed by a factor of 10 after $26$ and $33$ epochs.
Other implementation details are in Appendix~\ref{appendix:somethingv2}. 

\paragraph{Results.}
We investigate how well VideoMix improves the generalizability of action recognition models in the challenging benchmark beyond the Kinetics.
To separate the fine-tuning effects of VideoMix, it is applied only during the fine-tuning stage. The pretrained model is the same as the baseline. 
Table~\ref{table:experiment:something} shows the results. 
We observe that VideoMix improves the top-1 accuracies of SlowOnly-50 ($8\times8$) and SlowFast-50 ($8\times8$) by $+\mathbf{0.9}\%$ and $+\mathbf{0.8}\%$ against the baselines, respectively. 
VideoMix is effective on Something-Something-V2 as well.

\subsection{Weakly Supervised Temporal Action Localization}
\label{section:experiments:wstal}

The goal of weakly supervised temporal action localization (WSTAL) is to localize actions in untrimmed videos with a classifier trained using video-wise class labels only. 
Given a video input sequence, WSTAL model predicts the sequence's class label and also generates one-dimensional temporal proposals to localize actions in the video. 
WSTAL models do not exploit temporal action annotations during training and the generated temporal proposals are evaluated on validation videos with validation ground truth annotations. 
To localize the action instances well, a video model recognizes action categories from full video sequences and not focus on small discriminant frames of the action.
Through the WSTAL experiments, we verify that VideoMix improves the temporal localization ability of an action recognition models by guiding them to attend on wider frames of action. 
To evaluate the temporal localizability, we apply VideoMix over the baseline WSTAL methods.

\paragraph{Dataset.}
We conduct weakly supervised temporal action localization (WSTAL) task on THUMOS’14 dataset~\cite{THUMOS14}. 
THUMOS’14 dataset originally consists of 13,320 trimmed videos for training and 2,584 untrimmed videos for validation with 101 action categories.
We follow previous WSTAL methods' setting~\cite{lee2020bas,nguyen2018weakly,shou2018autoloc}.
We train WSTAL models with the 20 class subset of the untrimmed videos, which consists of 200 untrimmed videos without temporal annotations. 
The temporal localization performance of a model is evaluated by 212 untrimmed videos with temporal annotations. 
WSTAL on THUMOS’14 dataset is a challenging task since the length of untrimmed video could be quite long (up to 26 minute) and multiple actions could exist in a video. 

\paragraph{Training and inference.}
For training, we first extract I3D~\cite{carreira2017quo} features from the training videos as done in~\cite{lee2020bas}.
We sample $750$ video segments from a training video and RGB frames and optical flows are separately fed into dual-path I3D network. 
Each RGB and optical flow frame results in $1024$-dimensional feature, thus the dimension of extracted feature for a video is $750\times1024$ for RGB input, and $750\times2048$ for both using RGB and optical flow input (RGB+flow).
The WSTAL model, which consists of two $3\times3$ convolutional layers followed with LeakyReLU activation and a $1\times1$ convolutional layer, takes the extracted features and predicts its class label.
Since the network is trained with pre-computed features, we do not consider spatial VideoMix, but utilize \textbf{temporal VideoMix} on the extracted features.
Implementation details are in Appendix~\ref{appendix:wstal}.
For evaluation, temporal class activation mapping (T-CAM)~\cite{zhou2016CAM,nguyen2018weakly} is utilized to localize action instances along the temporal dimension. 
We threshold T-CAM below 50$\%$ of the highest value, and all 1-dimensional continuous segments are considered as action instance proposals as in~\cite{singh2017hide}. 
Evaluation metric is average precision (AP) of intersection over union (IoU) thresholds from $0.1$ to $0.9$, and we report their mean value (mAP) as in~\cite{singh2017hide,nguyen2018weakly,lee2020bas}.

\begin{table}[t]
\centering
\begin{tabular}{@{}lcc@{}}
\toprule
Method & mAP \\ \midrule
I3D RGB (Baseline) & 13.3 \\
+ Hide-and-Seek~\cite{singh2017hide} & 13.6 \\
+ Mixup~\cite{zhang2017mixup} & 14.0 \\
+ VideoMix & \textbf{14.2} \\ \midrule
I3D RGB+Flow (Baseline) & 17.8 \\
+ Hide-and-Seek~\cite{singh2017hide} & 18.2 \\
+ Mixup~\cite{zhang2017mixup} & 18.2 \\ 
+ VideoMix & \textbf{19.3} \\ \midrule
\end{tabular}
\caption{\textbf{Weakly supervised temporal action localization.} Results are on THUMOS14.
}
\label{table:experiment:wstal}
\end{table}

\paragraph{Results.}
Table~\ref{table:experiment:wstal} shows the WSTAL performances with and without optical flow features.
We compare VideoMix against the baselines including Hide-and-Seek~\cite{singh2017hide}, which has been reported to improve weakly supervised object localization on static images and WSTAL on videos.
We also compare against Mixup~\cite{zhang2017mixup}. 
We observe that VideoMix improves the temporal localization accuracy mAP of baseline by $+\mathbf{0.9}\%$ and $+\mathbf{1.5}\%$ with and without optical flows, respectively.
VideoMix also outperforms the Hide-and-Seek and Mixup in both scenarios showing its superior temporal localization ability.
We also conducted VideoMix with a stronger baseline, W-TALC~\cite{paul2018w}, and confirmed that VideoMix improves the performance of W-TALC from 31.1 to \textbf{32.3 (+1.2)} mAP, which is at the state-of-the-art level.

\subsection{AVA Action Detection}
\label{section:experiments:ava}

Kinetics pretraining is a widely-used practice for many video recognition tasks \cite{gu2018ava,sun2018actor,feichtenhofer2019slowfast}. 
We validate whether VideoMix pretrained models bring better performance on the downstream task of detecting actions in videos. 

\paragraph{Dataset.}
AVA v2.2 dataset~\cite{gu2018ava} consists of 235 training and 64 validation videos of human actions in videos. 
Each video is 15 minutes long, and the action locations are densely annotated as bounding boxes in space and time.
We follow the protocol in~\cite{gu2018ava} to train and evaluate the detection of 60 human action classes.
We use the mean average precision (mAP) metric to measure the performance of video action detection using a frame-level intersection-over-union (IoU) threshold $0.5$.

\paragraph{Detection framework.}
Our detector is based on the Faster R-CNN~\cite{ren2015faster} architecture, which is modified as in~\cite{feichtenhofer2019slowfast} to adapt to the video action detection task.
The spatial stride of the final convolutional block is reduced from $2$ to $1$ to increase the feature map size. The 2D RoIAlign layer~\cite{he2017mask} is replaced by the 3D RoIAlign. 
SlowOnly-50 ($8\times8$) and SlowFast-50 ($8\times8$) have been used as the backbone network for the detection framework. 
We use the human bounding box proposals provided by~\cite{feichtenhofer2019slowfast} computed by an off-the-shelf human detector fine-tuned on AVA persons. 

\paragraph{Training and inference.}

We initialize two detectors with the weights pretrained on Kinetics-400 with or without VidoeMix. We apply the same fine-tuning strategy afterwards to separate the effect of VideoMix on pretraining. 
We train detectors for $20$ epochs using the SGD optimizer with initial learning rate $0.1$ decayed by factor $0.1$ at $10$ and $15$ epoch.
The spatial dimension of the shorter side is resized to 256 pixels while maintaining the aspect ratio. 
$64$ consecutive frames are extracted for training.
Further details are in Appendix~\ref{appendix:ava}. 

\begin{table}[t]
\centering
\begin{tabular}{@{}lcc@{}}
\toprule
\multirow{2}{*}{Backbone}             & VideoMix & \multirow{2}{*}{val mAP} \\
 & Pretrained &  \\ \bottomrule
SlowOnly-50 ($8\times8$)        &       &  19.1 \\
SlowOnly-50 ($8\times8$)        &   \checkmark    &  \textbf{ 20.4  }  \\
\bottomrule
SlowFast-50 ($8\times8$)        &        &    23.2  \\
SlowFast-50 ($8\times8$)        &   \checkmark    & \textbf{ 24.9 } \\ \toprule
\end{tabular}
\caption{\textbf{AVA action detection.} Impact of VideoMix-pretraining on the transfer learning of AVA action detection.
}
\label{table:experiment:ava}
\end{table}

\paragraph{Results.}
Table~\ref{table:experiment:ava} shows the performances of our detector on the AVA benchmark. 
Pretraining the detector with VideoMix improves the performance of SlowOnly-50 and SlowFast-50 to $\mathbf{20.4}$ ($+\mathbf{1.3}$) and $\mathbf{24.9}$ ($+\mathbf{1.7}$) mAP, respectively. Switching the pretrained weights to the VideoMix version introduces the gain in detection performance for free. 
The weights will be published in the future.

\section{Conclusion}
\label{section:conclusion}

We have analyzed the augmentation strategies for video action classification task. We have introduced VideoMix, a simple, efficient, and effective augmentation method.
On Kinetics action recognition, VideoMix improves the top-1 accuracies of SlowOnly-50 and SlowFast-50 by $+1.3\%$ and $+0.5\%$, respectively. 
On Something-Something-V2 dataset, VideoMix brings $+0.9\%$ and $+0.8\%$ gains in top-1 accuracies on SlowOnly-50 and SlowFast-50, respectively. 
VideoMix also improves the localization ability of the classifiers: on weakly supervised temporal action localization (WSTAL), VideoMix consistently improves the localization accuracy over the baselines.
Finally, VideoMix improves the Kinetics-pretrained model for the transfer-learning task of video action detection. VideoMix, as well as VideoMix pretrained weights, provide a simple and cheap solution to boost up the video recognition performances across diverse tasks.

{\small
\section*{Acknowledgement}
We would like to thank NAVER AI LAB team members, especially Jung-Woo Ha for his helpful feedback and discussion.
}

{\small
\bibliographystyle{ieee_fullname}
\bibliography{egbib}
}

\clearpage

\appendix

\section{VideoMix Algorithm}
\label{appendix:algorithm}

\algnewcommand\Input{\item[\textbf{Input:}]}%
\algnewcommand\Output{\item[\textbf{Output:}]}%
\begin{algorithm*}[!t]
\small
  \caption{Pseudo-code of VideoMix}
  \begin{algorithmic}[1]
    \For {each iteration}
    \State input, target = get\_minibatch(dataset) 
    \Comment{input: $(\text{N}\times C \times T \times W\times H)$} %
    \Statex $ $ \Comment{target: $(\text{N}\times \#\text{Classes})~~~~~~~~~$} 
        \If{mode $==$ training}
            \State input$_\text{shuff}$, target$_\text{shuff}$ = shuffle\_minibatch(input, target) 
            \State $\lambda$ = $\text{Unif}(0,1)$
            \If{VideoMix\_mode $==$ Spatial} 
                \State $w_c$, $w_h$ = $\text{Unif}(0,W)$, $\text{Unif}(0,H)$
                \State $w_1$, $w_2$ = $\text{clip}(w_c-W\frac{\sqrt{\lambda}}{2})$, $\text{clip}(w_c+W\frac{\sqrt{\lambda}}{2})$
                \State $h_1$, $h_2$ = $\text{clip}(h_c-H\frac{\sqrt{\lambda}}{2})$, $\text{clip}(h_c+H\frac{\sqrt{\lambda}}{2})$
                \State $t_1$, $t_2$ = $0$, $T$
            \ElsIf {VideoMix\_mode $==$ Temporal}
                \State $t_c$ = $\text{Unif}(0,T)$
                \State $w_1$, $w_2$ = $0$, $W$
                \State $h_1$, $h_2$ = $0$, $H$
                \State $t_1$, $t_2$ = $\text{clip}(t_c-T\frac{{\lambda}}{2})$, $\text{clip}(t_c+T\frac{{\lambda}}{2})$
            \ElsIf {VideoMix\_mode $==$ Spatio-Temporal}
                \State $w_c$, $w_h$, $t_c$ = $\text{Unif}(0,W)$, $\text{Unif}(0,H)$, $\text{Unif}(0,T)$
                \State $w_1$, $w_2$ = $\text{clip}(w_c-W\frac{\sqrt[3]{\lambda}}{2})$, $\text{clip}(w_c+W\frac{\sqrt[3]{\lambda}}{2})$
                \State $h_1$, $h_2$ = $\text{clip}(h_c-H\frac{\sqrt[3]{\lambda}}{2})$, $\text{clip}(h_c+H\frac{\sqrt[3]{\lambda}}{2})$
                \State $t_1$, $t_2$ = $\text{clip}(t_c-T\frac{\sqrt[3]{\lambda}}{2})$, $\text{clip}(t_c+T\frac{\sqrt[3]{\lambda}}{2})$                
            \EndIf
            
            \State input[:, :, $t_1$:$t_2$, $w_1$:$w_2$, $h_1$:$h_2$] = input$_\text{shuff}$[:, :, $t_1$:$t_2$, $w_1$:$w_2$, $h_1$:$h_2$]
            \State $\lambda$ = $\frac{(t2-t1)\times(w2-w1)\times(h2-h1)}{(T\times W\times H)}$ \Comment{Adjust lambda to the exact fraction ratio.}
            \State target = $\lambda$ * target + (1 - $\lambda$) * target$_\text{shuff}$%
        \EndIf
        \State output = model\_forward(input)
        \State loss = compute\_loss(output, target)
        \State update\_model(model, loss)
    \EndFor
  \end{algorithmic}
  \label{alg:videomix_algorithm}
\end{algorithm*}

We describe code-level algorithm of VideoMix variants in Algorithm~\ref{alg:videomix_algorithm}.
The input video of a minibatch is $(N\times C \times T \times W \times H)$-size tensor, where $N$, $C$, $T$, $W$, and $H$ denote the size of a minibatch, the channel size, the the width, and the height of a frame.
VideoMix first shuffles the order of the minibatch along the first dimension of the tensor. 
Next $\lambda$ is sampled from the uniform distribution.
Then the cuboid coordinate $\mathbf{C}=(t_1,t_2,w_1,w_2,h_1,h_2)$ are sampled corresponding to the type of VideoMix. 
Note that `clip' function truncates the coordinates to fit in the frame space (e.g., $\text{clip}(w)$=$\text{min}(\text{max}(w,W),0)$). 

\textbf{Spatial VideoMix} ({line 7-10}) is the same as we described in the main paper. 
For \textbf{Temporal VideoMix} (line 12-15), we only samples $t_1$ and $t_2$.
For \textbf{Spatio-temporal Videomix} (line 17-20), $t_1$, $t_2$, $w_1$, $w_2$, $h_1$, and $h_2$ are simultaneously sampled. 
After sample a cuboid $\mathbf{C}$, we combine two videos by cutting and inserting the cuboid region, and $\lambda$ is adjust by computing the exact fraction ratio of the cuboid. 
The target label is also blended by interpolation manner. 
Note that VideoMix is simple and easy to implement with few lines, but it is very effective on various video tasks.

\section{Additional Experiments}
\label{appendix:more_baseline}

\subsection{HMDB-51 and UCF-101}

We evaluated VideoMix on HMDB-51~\cite{kuehne2011hmdb} and UCF-101~\cite{soomro2012ucf101} benchmarks. Table~\ref{hmdb51} and Table\ref{ucf101} shows the results.
Our method consistently boosts the top-1 accuracy against the baseline models.

\begin{table}[h]
\centering
\begin{tabular}{@{}ll@{}}
\toprule
Methods              & top1 acc.  \\ \midrule
I3D (Baseline)	& 66.0 \\
I3D + VideoMix	&\textbf{66.9 (+0.9)}   \\ \midrule
\end{tabular}
\caption{HMDB-51 benchmark results.}
\label{hmdb51}
\end{table}

\begin{table}[h]
\centering
\begin{tabular}{@{}ll@{}}
\toprule
Methods              & top1 acc.   \\ \midrule
VGG16 (Baseline)	& 79.8 \\
VGG16 + VideoMix & \textbf{81.7 (+2.1)}\\ 
I3D (Baseline)	& 93.3 \\
I3D + VideoMix	& \textbf{93.4 (+0.1)} \\ \midrule
\end{tabular}
\caption{UCF-101 benchmark results.}
\label{ucf101}
\end{table}

\subsection{Additional ablation study}

We subsample the mini-Kinetics dataset (10$\%$, 25$\%$, 50$\%$, and 100$\%$) and train the SlowOnly-34 model on the sampled datasets. 
Table~\ref{ablation:datasize} shows the results. 
Results show that VideoMix consistently improves accuracies against the baseline for all the subsets of the Mini-Kinetics dataset.

\begin{table}[h]
\centering
\begin{tabular}{@{}lcccc@{}}
\toprule
Dataset size &	10$\%$	& 25$\%$	& 50$\%$	& 100$\%$ \\ \midrule
Baseline	& 31.2 &	47.2	& 67.6 & 	75.2 \\
VideoMix	& \textbf{33.3}	& \textbf{49.6}	& \textbf{68.0}	& \textbf{77.6} \\ \midrule 
\end{tabular}
\caption{\textbf{Impact of dataset sizes.} Top-1 accuracy is reported.}
\label{ablation:datasize}
\end{table}

To see the complementarity among the data augmentation methods (e.g., Cutout, Mixup, RandAug, and VideoMix), we conduct various combinations of data augmentation methods.
Starting from the standard augmentation (flip and random resize), we stack up VideoMix, Mixup, Cutout, and RandAugment.
Table~\ref{ablation:adding} shows the results. 
VideoMix boosts performance against the standard augmentation, a relatively weak augmentation. 
Combining VideoMix with other strong augmentations (Cutout or RandAugment) degrades performance since the combination leads to an excessive amount of regularization.
Mixup also shows similar tendency that combining with other augmentations leads to degraded performance. 

\begin{table}[h]
\tabcolsep=0.05cm
\centering
\begin{tabular}{@{}cccccc@{}}
\toprule
Standard Aug. &	VideoMix	& Mixup & Cutout	& RandAug	& top1  \\ \midrule
\checkmark	& & & &	&		75.2 \\
\checkmark	& \checkmark & &	& &		\textbf{77.6} \\
\checkmark	& \checkmark & & \checkmark & &		76.3 \\
\checkmark	& \checkmark & &	& \checkmark &	76.4 \\ 
\checkmark	& \checkmark & & \checkmark & \checkmark &	73.5 \\
\checkmark	& & \checkmark &  &	&	77.0 \\
\checkmark	& & \checkmark & \checkmark &	&	74.2 \\
\checkmark	& & \checkmark & \checkmark & \checkmark &	76.7  \\
 \midrule
\end{tabular}
\caption{\textbf{Complementarity among data augmentations.}}
\label{ablation:adding}
\end{table}
\section{Something-Something-V2 Action Recognition}
\label{appendix:somethingv2}
We describe the implementation details for Something-Something-V2 Action Recognition here. 

We train models on Something-Something-V2 dataset for 40 epochs with the batch size 64 and learning rate 0.01.
The learning rate is decayed by a factor of 10 after 26 and 33 epochs. 
Other training details are the same as Kinetics (Section 4.1 of the main paper), except for the standard data augmentation. 

The standard data augmentation of Kinetics experiments is that horizontal flipping, random cropping, and temporal uniform sampling. 
Temporal uniform sampling samples a random clip of the entire sequences with a uniform frame interval. 

For Something-Something-V2, we do not use horizontal flipping augmentation since the action's direction is critical for this dataset (e.g., there is  a `pushing something from left to right' action category). 
Also, we sample frames with temporally perturbed interval instead of temporal uniform sampling. 
In detail, we first split the entire frames with $T$ bins ($T$ is the number of sampled frames), and we select a frame from each bin and aggregate $T$ frames.

For inference, we use 3 spatial crops and single temporal crop.
\section{THUMOS'14 Weakly Supervised Temporal Action Localization}
\label{appendix:wstal}

We describe the implementation details for THUMOS'14 Weakly Supervised Temporal Action Localization (WSTAL) task. 

We utilize the codebase\footnote{https://github.com/Pilhyeon/BaSNet-pytorch} of \cite{lee2020bas} for I3D~\cite{carreira2017quo} baseline.
We extract I3D~\cite{carreira2017quo} features from training video using this repository\footnote{https://github.com/piergiaj/pytorch-i3d}.
We sample $750$ video segments from a training video.
Each segment has $16$ frames and the segments are not overlapped. 

The input video has RGB frames and also optical flows, and they are separately fed into dual-path I3D network. 
Each segment of RGB and optical flow frames results in 1024-dimensional feature.
Thus the dimension of extracted feature for a video (i.e., $750$ segments) is $750\times1024$ for RGB input, and $750\times2048$ for both using RGB and optical flow input (RGB+flow).
We apply \textbf{Temporal VideoMix} along the temporal dimension (i.e., the first axis of the feature $750\times2048$) on the extracted feature.

To see the effectiveness of VideoMix with more stronger baseline, we conducted VideoMix with W-TALC~\cite{paul2018w} using the official pytorch codebase\footnote{https://github.com/sujoyp/wtalc-pytorch}. 
We follow the original codebase's setting and we apply \textbf{Temporal VideoMix} along the temporal dimension as in the I3D experiment.

\section{AVA Action Detection}
\label{appendix:ava}

Our action detector is based on the Faster R-CNN~\cite{ren2015faster} architecture, which is modified as in~\cite{feichtenhofer2019slowfast} to adapt to the video action detection task. 
We use PySlowFast\footnote{https://github.com/facebookresearch/SlowFast} and Detectron2\footnote{https://github.com/facebookresearch/detectron2} codebases.
The spatial stride of the final convolutional block is reduced from 2 to 1
to increase the feature map size.
We extend 2D RoIAlign layer~\cite{he2017mask} to 3D RoIAlign layer, which extracts RoI features spatially and then aggregate via global average pooling. 
We use the human bounding box proposals provided by~\cite{feichtenhofer2019slowfast} computed by an off-the-shelf human detector fine-tuned on AVA persons, which is a Faster RCNN with a ResNeXt-101-FPN~\cite{xie2017resnext} backbone.
The person region proposals are detected by human detector with a confidence threshold of $0.8$.

We train detectors for 20 epochs using the SGD optimizer with initial learning rate 0.1 decayed by factor 0.1 at 10 and 15 epoch. 
The spatial size of the input video is $224\times224$, and $64$ consecutive frames are extracted for training.
For inference, the spatial dimension of the shorter side is resized to $256$ pixels while maintaining the aspect ratio. 

\section{Spatio-temporal Class Activation Mapping}
\label{appendix:st-cam}

We describe the spatio-temporal class activation mapping (ST-CAM) which is extended from spatial CAM of the original paper~\cite{zhou2016CAM}.
We use a SlowOnly-50 ($8\times8$) network~\cite{feichtenhofer2019slowfast} which is pretrained on Kinetics-400~\cite{kinetics} dataset. 

To obtain a ST-CAM of a given video input, we first extract the final feature map of the SlowOnly-50 network before the global average pooling layer. 
The temporal dimension of the feature map is $8$ which is the same as the number of input frames.
We reduce the spatial stride of the last convolution from $2$ to $1$ of the original SlowOnly-50 network, so that the spatial dimension of the feature map is $14 \times 14$ to clearly see the CAM heatmap, while the original size is $7 \times 7$.
The number of channels of the feature map is $2048$.
Then, similar to the original paper~\cite{zhou2016CAM}, the extracted final feature map $(2048\times8\times14\times14)$ is multiplied with the fully-connected layer's weight corresponding to the target class $(2048\times1)$, resulting in $(8\times14\times14)$-dimensional spatio-temporal class activation mapping.
We sub-sampled $4$ frames of the ST-CAM in the main paper due to the limit of page width. 

\begin{figure*}[!t]
\centering
\includegraphics[width=0.77\linewidth]{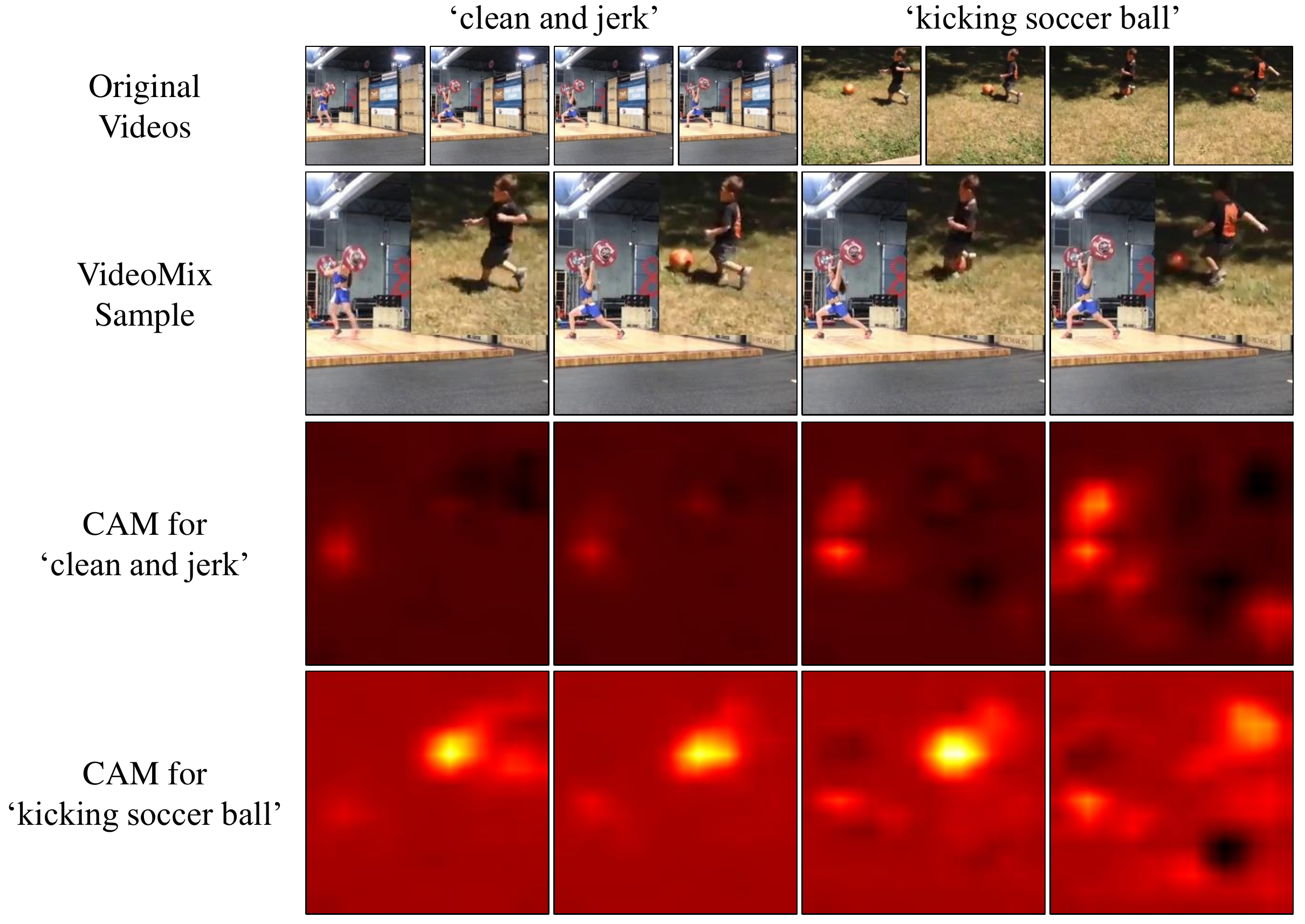}
\caption{Spatio-temporal class activation mapping (ST-CAM) on VideoMix sample of the ``clean and jerk'' and ``kicking soccer ball'' videos.}
\label{supp:cam1}
\end{figure*}

\begin{figure*}[!t]
\centering
\includegraphics[width=0.77\linewidth]{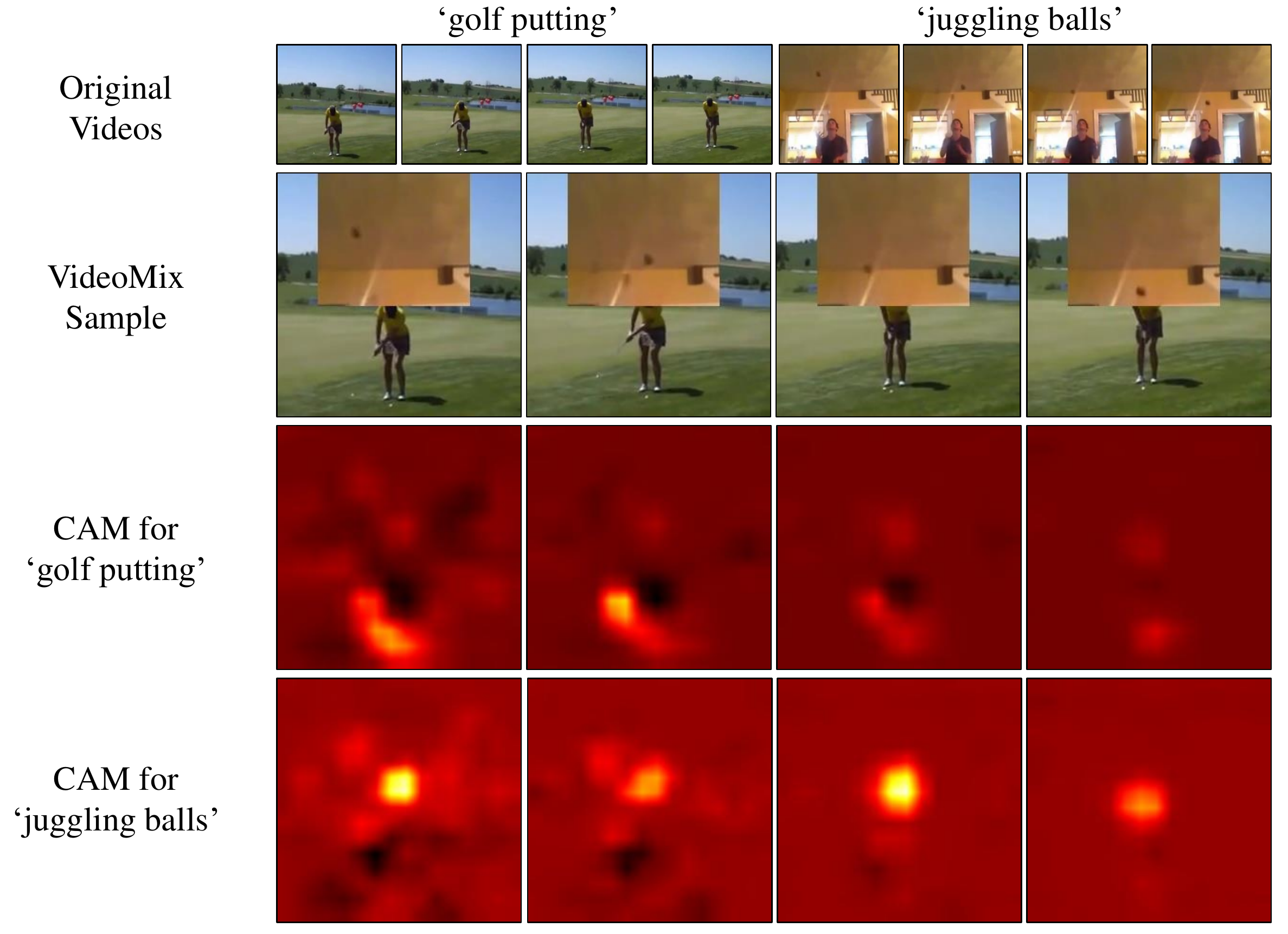}
\caption{Spatio-temporal class activation mapping (ST-CAM) on VideoMix sample of the ``golf putting'' and ``juggling balls'' videos.}
\label{supp:cam2}
\end{figure*}

\begin{figure*}[!t]
\centering
\includegraphics[width=0.77\linewidth]{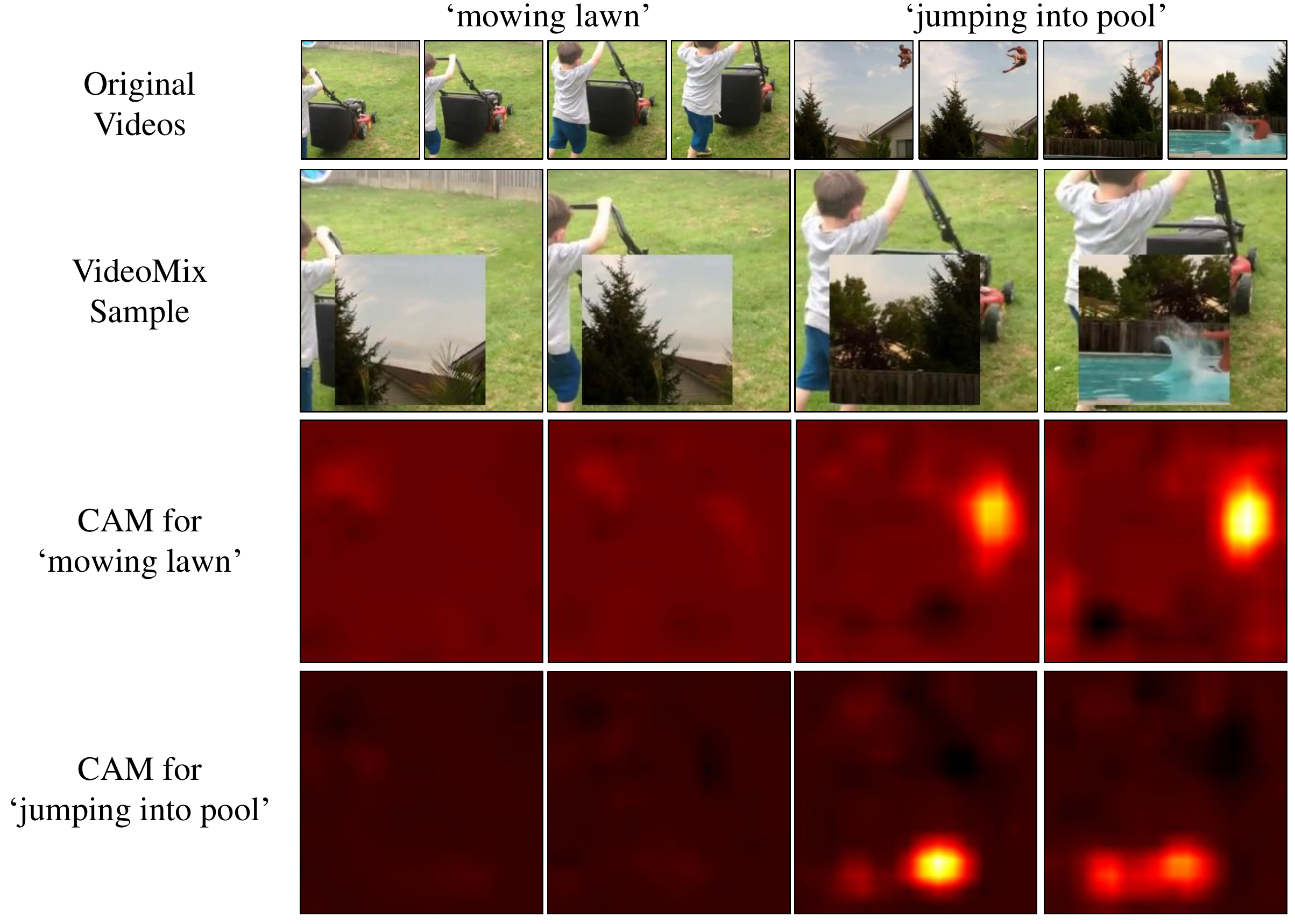}
\caption{Spatio-temporal class activation mapping (ST-CAM) on VideoMix sample of the ``mowing lawn'' and ``jumping into pool'' videos.}
\label{supp:cam3}
\end{figure*}

\begin{figure*}[!t]
\centering
\includegraphics[width=0.77\linewidth]{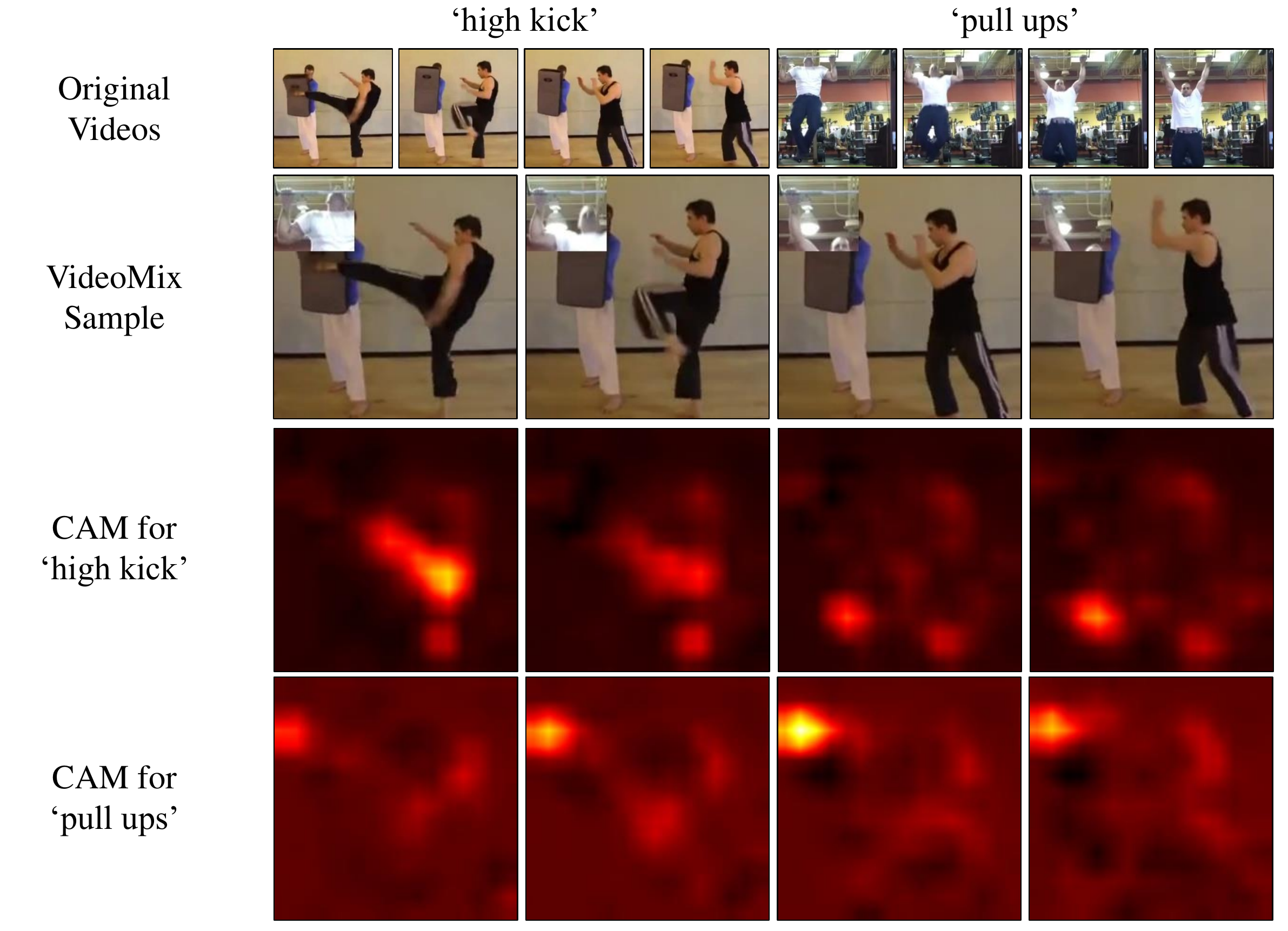}
\caption{Spatio-temporal class activation mapping (ST-CAM) on VideoMix sample of the ``high kick'' and ``pull ups'' videos.}
\label{supp:cam4}
\end{figure*}

We present additional ST-CAM visualizations in Figure~\ref{supp:cam1},~\ref{supp:cam2},~\ref{supp:cam3}, and~\ref{supp:cam4} which show the VideoMix samples and corresponding class activation maps with respect to the two action classes.


\end{document}